\documentclass[a4paper]{article}

\usepackage{INTERSPEECH2018}
\usepackage{amsmath,graphicx}

\usepackage{pgfplots}
\usepackage{tikz}

\pgfplotsset{compat=1.14}

\title{Subword and Crossword Units for CTC Acoustic Models}
\name{Thomas Zenkel$^1$, Ramon Sanabria$^2$, Florian Metze$^2$ and Alex Waibel$^{1,2}$}
\address{
  $^1$Karlsruhe Institute of Technology, Karlsruhe, Germany\\
  $^2$Carnegie Mellon University, Pittsburgh, PA, U.S.A.}
\email{thomas.zenkel@kit.edu, ramons@cs.cmu.edu, fmetze@cs.cmu.edu, ahw@cs.cmu.edu}

\begin{document}

\maketitle
\begin{abstract}
This paper proposes a novel approach to create a unit set for CTC-based speech recognition systems. By using Byte-Pair Encoding we learn a unit set of arbitrary size on a given training text. In contrast to using characters or words as units, this allows us to find a good trade-off between the size of our unit set and the available training data. We investigate both crossword units, which may span multiple words, and subword units. By evaluating these unit sets with decoding methods using a separate language model, we are able to show improvements over a purely character-based unit set. 
\end{abstract}

\section{Introduction}
\label{sec:intro}
Traditional automatic speech recognition (ASR) systems consist of three components: a phoneme-based acoustic model (AM), a word-based language model (LM), and a pronunciation lexicon, which maps a sequence of phonemes to words~\cite{rabiner1989tutorial}.
This setup works well for systems based on hidden Markov models (HMMs) as well as for `end-to-end' systems~\cite{graves2014towards, chan2016listen}, where the AM is trained towards a sequence loss, such as Connectionist Temporal Classification (CTC) \cite{graves2006connectionist}. An expert-created phone dictionary will contain accurate pronunciations, and often multiple variants for frequent words, which facilitates the AM's task.

A CTC model's `spike-train' pattern marginalizes over all possible alignments of the output symbols, and bi-directional long-short term memory (LSTM) networks can learn temporal patterns well~\cite{graves2005bidirectional}. CTC models thus perform surprisingly well, even when using context-independent characters as the AM's units, and in languages with `irregular' pronunciations, such as English. Because of CTC's independence assumption, the model cannot however explicitly learn co-articulation patterns, which should help improve performance.

If characters are used as the units of the AM, the LM can directly be implemented as a recurrent neural network (RNN)~\cite{maas2015lexicon}. Together with a CTC AM, it allows the creation of `all-neural' systems, which are not restricted to decoding within the weighted Finite State Transducer (WFST) framework.

Recent work shows that CTC models can directly predict word units, albeit on an extremely large training corpus~\cite{soltau2016neural}.
\cite{audhkhasi2017direct}~circumvent this problem by pre-training the AM with phoneme sequences. Unfortunately, using words as output units re-introduces a fixed vocabulary and only frequent words will be trained robustly. 

Character n-gram units provide a trade-off between character and word-based unit sets. When using n-grams, the number of possible outputs increases exponentially with respect to the longest n-gram length. For this reason \cite{chan2016latent, liu2017gram} constrain the length of their n-gram units to a fixed length and only use selected subword n-grams. 


Additionally, none of the mentioned grapheme-based approaches deal with crossword pronunciation phenomena like contractions and reductions. In conversational English, `kind of' and `going to' are often pronounced as `kinda' and `gonna.' It may be useful to treat such phenomena as their own units, since the AM could then learn these reduced pronunciations \cite{finke1997speaking}.

In this work we propose a method to create a unit set for CTC AMs based on the Byte-Pair Encoding (BPE) algorithm \cite{gage1994new}. This tackles two shortcomings of n-gram units: we do not restrict our units to have a fixed length, and we only create units that frequently appear in a given training corpus. Due to the iterative construction of the unit set, it is easy to empirically determine the best trade-off between the size of our unit set and the number of training labels. 
Additionally, we do not restrict our units to be part of a word only. In this paper we investigate the use of subword and crossword BPE units for CTC speech recognition systems and compare them to character and word units. We compare WFST and RNN decoding approaches. 

Compared to other approaches to learning the units of an ASR system, the main advantage of this method is that it does not rely on phonetics at all, and can be trained on text only, which greatly speeds up system-building efforts. Additionally, we still keep the ability of recognizing arbitrary words and do not constrain our system by using a fixed vocabulary.

\section{Related Work}
\label{sec:relatedWork}
The problem of selecting a unit set is closely related to finding a decomposition of the target sequence into basic units. The straightforward variant is to rely on a fixed decomposition of the target sequence, hence a given target sequence will always be split into the same units. 

Many approaches using fixed decomposition have been proposed to solve different problems. An approach used for language models is to keep frequent words as units and split the remaining units into syllables \cite{mikolov2012subword}. In machine translation systems the Byte-Pair Encoding algorithm is used to split infrequent words into subword units \cite{sennrich2015neural}.  This can improve the translation quality of compounds as well as cognates and loanwords. For speech recognition applications, \cite{finke1997speaking} deals with crossword pronunciation phenomena. By adding frequent multiwords like `sort of' and `kind of' and their corresponding pronunciations to the pronunciation lexicon they were able to improve recognition performance. 

Instead of using a fixed decomposition, it is also possible to learn a variable decomposition of the target sequence. The decomposition can for example depend on  speaking style for speech recognition tasks. This approach was successfully applied to neural speech recognition systems \cite{chan2016latent, liu2017gram}. \cite{liu2017gram} extends the CTC loss function to learn the alignment between the input and the target sequence as well as a suitable decomposition of the target sequence. The unit set consists of character n-grams up to a fixed length and infrequent n-grams are omitted in the unit set. Because of the additional task of learning a decomposition, more training data is necessary.

Another recent trend is to combine word and character-based acoustic models. Word-based models use a finite vocabulary and model words which do not appear in the vocabulary as a special output token, normally called the OOV token. Instead of outputting the OOV token, \cite{li2018advancing, audhkhasi2017building} output characters from a different acoustic component for OOVs. 


\section{Unit Selection}
\label{sec:unitSelection}
For creating our units we use Byte-Pair Encoding (BPE) \cite{gage1994new}. BPE is a compression algorithm that iteratively replaces the most frequent pair of units (or bytes) with an unused unit. Our initial unit set for all our experiments consists of alphanumerical characters, tokens to represent various noises, and some special characters (-'/\&). With each step the unit set grows by one and thus we can create an arbitrary number of units. For example, if the units 'AB' and 'CDE' are already in the unit set and 'AB' frequently appears before 'CDE', the new unit 'ABCDE' will be created and added to the unit set. We use the BPE algorithm to create subword units as well as crossword units.

\subsection{Subword Units}
To create subword units we use the method proposed for machine translation in \cite{sennrich2015neural}. We start the algorithm using our initial unit set. A special token (in our case `@') is used to denote if the unit appears within a word. We do not use a dedicated space character, as the word boundaries are defined by the absence of the token `@' within a unit. Crossword boundaries are not considered in this case; we only count the co-occurrences of units within a word. Since BPE creates a new unit at each iteration, very frequent words are eventually merged into a single unit.

\subsection{Crossword Units}
To create crossword units we slightly change the algorithm used for the subword units. In place of an end of word symbol, we mark the beginning of each word with a capital letter. This idea is inspired by the unit set used for the speech recognition system in \cite{zweig2016advances}. For example, the utterance `i don't know' will be preprocessed to `IDon'tKnow'. Notice that we do not use a dedicated space character, since the word boundaries are modeled using capital letters. We now apply BPE and also merge across word boundaries. We argue that this method is superior to directly applying BPE to utterances containing spaces. When doing this, it is possible that the unit set will now include `know', `\_know', `know\_' as well as `\_know\_' (for visibility spaces are replaced by '\_').

\subsection{Comparison}
We compare the decomposition of a given target sequence in Table \ref{tab:split}. We show an example of both small and large unit sets for each approach, subword and crossword. The small set is created by applying 300 merge operations, while the large set is created by applying 10,000 merge operations. For small subword units, all words except very frequently appearing words are split into subword units. Large subword models almost resemble word units; however, very infrequent words or words that did not appear in the training corpus are still split into multiple subword units.

In contrast, the crossword units also include units consisting of multiple words. Small subword units already contain multi-words for very frequent expressions. The units can become longer for the large model and also less frequent expressions are merged into a single unit. The length of both the subword and the crossword sequences is significantly lower than the length of the character sequences.

\begin{table}[th]
  \centering
  \caption{An example utterance split into characters and processed with the subword and crossword BPE algorithm when creating 300 and 10,000 additional units, respectively. In the subword model the character '@' denotes that the unit is not the end of the word.}
  \begin{tabular}{ c  p{5.6cm} }
  Method & Utterance \\
  \hline
  Original & you know it's no not even cold weather \\
  Character & y o u k n o w i t ' s n o n o t e v e n c o l d w e a t h e r \\
  Subword 300 & you know it's no not even co@ ld w@ ea@ ther \\
  Subword 10k & you know it's no not even cold weather \\
  Crossword 300 & YouKnow It's No Not E ven Co ld We at her \\
  Crossword 10k & YouKnowIt's No NotEven \mbox{ColdWeather} \\
\end{tabular}
\label{tab:split}
\end{table}

Since we always keep single characters in the unit set, we are able to model arbitrary words with both approaches. This enables our speech recognition system to be open vocabulary.

\section{Acoustic Model}
\label{sec:acousticModel}

The AM of our system is composed of four bidirectional LSTM layers \cite{hochreiter1997long} with 320 units in each direction followed by a softmax layer. The size of the softmax layer depends on the unit set we use. We jointly train the whole model under the CTC loss function~\cite{graves2006connectionist}.

To train the AM we use the 300h Switchboard data set (LDC97S62). We perform data augmentation as in \cite{zenkel2017}, creating 3 subsamples with a reduced frame rate (\textit{i.e.} from 10ms to 30ms) from each original sample.
The code is open-sourced in EESEN \cite{miao2015eesen}; the \texttt{tf\_clean} branch was used to train our models. To improve training stability, we pre-trained each model using only character labels until its convergence. Afterwards, we train  with one of the unit sets described in section \ref{sec:unitSelection}. We optimize the parameters of the network using stochastic gradient descent. We halve the learning rate after each epoch where the validation accuracy does not improve.

\section{Decoding Strategies}
\label{sec:decodingStrategies}
In this section we briefly summarize the different approaches to generate a transcription given the static sequence of probabilities generated by the acoustic model. For a more detailed description we refer readers to \cite{zenkel2017}. 

\subsection{Greedy Decoding}\label{decodinggreedy}
To estimate the quality of the AM we perform a greedy search without adding any linguistic knowledge by selecting the most probable unit at each frame \cite{graves2006connectionist}. By removing repeated units and the blank token, we are able to create a string of units. For the subword units, we insert a space to seperate words between each decoded unit (e.g. `oh ye@ ah') and remove the sequence \\ `@ ' to get a sequence of words (e.g. `oh yeah'). In the crossword case, we simply concatenate the decoded units (e.g. `OhYeah') and replace all uppercase character with their lowercase counterpart and a space (e.g. `oh yeah'). 

\subsection{Weighted Finite State Transducer}\label{wfstdef}
To improve over simple greedy search, the Weighted Finite State Transducer (WFST) approach adds linguistic information at the word level \cite{miao2015eesen}. 

The search graph of the WFST is composed of a token WFST, which applies the CTC squash function (i.e. removing repeated characters and the blank token), a lexicon WFST that maps a sequence of units to words, and a grammar WFST that is modeled by a word based n-gram language model. 

The search graph is used to find the most probable word sequence. We use a trigram and a 4-gram LM smoothed with Kneser Ney discounting. We train the LMs with cleaned Switchboard and Fisher transcripts. We do not use WFST decoding for crossword units. However, this would be possible by introducing multi-words into the lexicon.

\subsection{Beam Search with a RNN Language Model}
In contrast to the WFST decoding, we combine the information of the AM and the RNNLM directly at each frame. This is possible because we train the LM on the same unit set as the AM. We additionally add symbols denoting the start and the end of a sequence to the LM unit set. To find the most probable output sequence, we apply beam search similarly to \cite{hwang2016character}. For a mathematical formulation of the search, we refer to \cite{zenkel2017}.

We use a two-layer LSTM network with 1024 hidden units at each layer and a 256-dimensional embedding layer as our LM. We use the same cleaned Switchboard and Fisher transcripts as our training corpus for the RNNLM. We optimize the network with adam \cite{kingma2014adam}. We halve the learning rate and restart adam whenever our validation cost does not decrease \cite{denkowski2017stronger}.

\section{Results}
\label{sec:experiments}
The 2000 HUB5 `Eval2000' (LDC2002S09) set is used for evaluation. The corpus consists of English telephone conversations and is divided into the `Switchboard' subset, which more closely resembles the training data, and the `Callhome' subset. We evaluate both the crossword and subword models with both beam search using an RNNLM and greedy decoding, which does not use additional linguistic knowledge from the LM. For subword models we also apply WFST decoding.
For each approach, we evaluate with 300, 600, 1k, 3k, 6k and 10k units.

\subsection{Evaluation of Subword and Crossword Units}
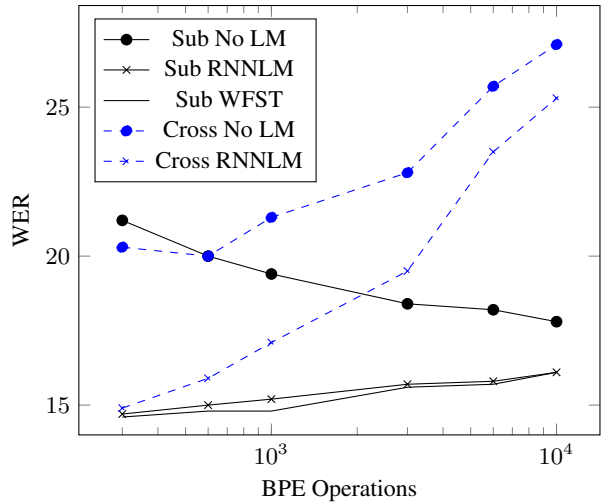
\begin{figure}
\begin{tikzpicture}
	\begin{axis}[
		xlabel=BPE Operations,
        ylabel=WER,
        ymin=14,
        xmode = log,
        legend pos=north west,
]

	\addplot[color=black,mark=*] coordinates {
		(300,21.2)
		(600,20.0)
		(1000,19.4)
		(3000,18.4)
        (6000, 18.2)
		(10000,17.8)
   };  
   
	\addplot[color=black,mark=x] coordinates {
		(300,14.7)
		(600,15.0)
		(1000,15.2)
		(3000,15.7)
        (6000, 15.8)
		(10000,16.1)
	};
    
	\addplot[color=black] coordinates {
		(300,14.6)
		(600,14.8)
		(1000,14.8)
		(3000,15.6)
        (6000, 15.7)
		(10000,16.1)
	};
    
	\addplot[color=blue,dashed, mark=*] coordinates {
		(300,20.3)
		(600,20.0)
		(1000,21.3)
		(3000,22.8)
        (6000,25.7)
		(10000,27.1)
   };
   
	\addplot[color=blue,dashed, mark=x] coordinates {
		(300,14.9)
		(600,15.9)
		(1000,17.1)
		(3000,19.5)
        (6000,23.5)
		(10000,25.3)
   };

\legend{Sub No LM, Sub RNNLM, Sub WFST, Cross No LM, Cross RNNLM}
\end{axis}
\end{tikzpicture}
\caption{Comparison of Subword (Sub) and Crossword (Cross) unit sets using no LM, a word-based LM (WFST), and an RNNLM. We report the Word Error Rate on the Switchboard subset of Eval2000.} 
\label{fig:plot}
\end{figure}

Figure \ref{fig:plot} summarizes the results. For the smallest unit set (300), the crossword model outperforms the subword model. While the word error rate (WER) of the subword model constantly decreases, the WER of the crossword model increases. One problem of the crossword model is that the length of the units constantly increases (e.g. ``You'reNotGoingTo''), while for the subword model units cannot be longer than a single word. We argue that it is more difficult for the AM to recognize long expressions. The other drawback of a bigger unit set is that we have fewer training examples per unit. This holds for both approaches; however, with the subword model frequent words will still have a higher number of training examples, as words cannot be represented as longer units.
As an example, consider the word `\textit{kinds}': in the subword 10k model, this unit has 539 training examples, while for the crossword 10k model, these are scattered between multiple units (``AllKindsOf'' (203), ``KindsOf'' (158), ``KindsOfThings'' (75) etc). Because of a lower number of training examples, we argue that the crossword model is not able to learn robust representations for these units. This is also supported by the fact that large crossword models output blanks in place of a reasonable unit in many situations. The deletion rate consistently increases with the size of the unit set, reaching 14.3\% for the crossword 10k model.


Adding linguistic knowledge during decoding always improves our results. For the large crossword models, the RNNLM is not able to fix the discussed drawbacks; even with a tuned insertion bonus, the deletion rate remains high (Table \ref{tab:errors}). We achieve the best results when using models with small unit sets. We argue that by using smaller units we can include the linguistic knowledge earlier in the search process. We do not have to wait until the AM has recognized a word, but in most cases can combine the information from the AM and the LM already at the subword level. 

\begin{table}[th]
   \caption{Substitution (S), Insertion (I), Deletion (D) and Word Error Rates (WER) for different unit sets using an RNNLM during decoding on the Switchboard subset of Eval2000.}
  \centering 
  \begin{tabular}{ c c c c c}%
  Method & S & D & I & WER \\%
  \hline
  Subword 300 & 8.8\% & 3.5\% & 2.4\% & 14.7\% \\
  Subword 10k & 8.0\% & 6.0\% & 2.1\% & 16.1\% \\
  Crossword 300 & 8.8\% & 4.0\% & 2.1\% & 14.9\% \\
  Crossword 10k & 9.9\% & 12.2\% & 3.2\% & 25.3\% \\
\end{tabular}
  \label{tab:errors}
\end{table}


We also found that the training time of each model varies according to the number of units used. For instance, a training epoch of the subword AM with 300 output units, 4 layers, and 320 cells takes 42 minutes, and the same model with 10k output units takes 166 minutes using an Nvidia GeForce GTX 1080 Ti. This is most likely due to the final projection layer which maps the hidden units towards a desired number of units.

\subsection{Recognition of unseen words}
While the RNNLM is slightly inferior to WFST decoding, it is still able to output arbitrary words and is not restricted by a fixed vocabulary. In Figure \ref{tab:errors}, we analyze the number of words which did not appear in the training set of the acoustic model but were nonetheless correctly recognized in the test set. We report these numbers for greedy decoding, which means without adding any LM information. We notice that we are able to recognize more previously unseen words when using a smaller unit set. With an increasing unit set size, fewer unseen words are recognized correctly. This situation also holds for decoding with the RNNLM. When using the same unit set for both the AM and the LM, we are still able to output arbitrary words.

\begin{figure}
\begin{tikzpicture}
	\begin{axis}[
		xlabel=BPE Operations,
        ymin=0,
        ymax=35,
        xmode = log,
        legend pos=north east,
]
	\addplot[color=black,mark=*] coordinates {
		(300,21)
		(600,17)
		(1000,16)
		(3000,8)
        (6000,4)
		(10000,2)
   };  
   
	\addplot[color=black,mark=x] coordinates {
		(300,25)
		(600,25)
		(1000,28)
		(3000,16)
        (6000,16)
		(10000,6)
	};
   
	\addplot[color=blue,dashed, mark=*] coordinates {
		(300,16)
		(600,18)
		(1000,11)
		(3000,5)
        (6000,5)
		(10000,3)
	};
    
	\addplot[color=blue,dashed, mark=x] coordinates {
		(300,24)
		(600,22)
		(1000,20)
		(3000,19)
        (6000,14)
		(10000,10)
   };

\legend{Sub No LM, Sub RNNLM, Cross No LM, Cross RNNLM}
\end{axis}
\end{tikzpicture}
\caption{Number of words which were correctly recognized during greedy and RNNLM decoding (y-axis), but did not appear in the acoustic training corpus.} 
\label{fig:plotOOV}
\end{figure}
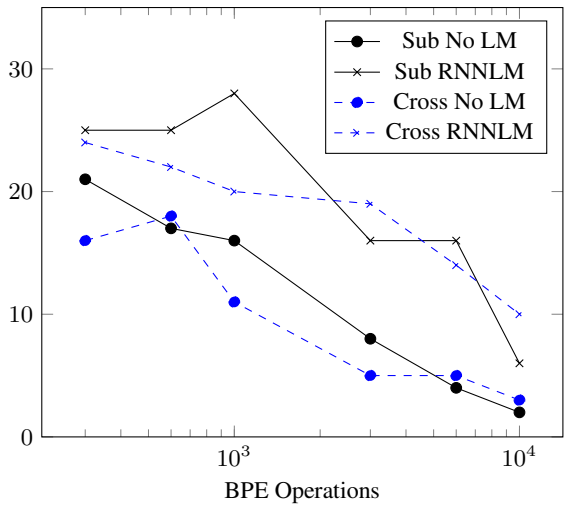

Most new words we are now able to recognize consist of previously seen words with different prefixes or suffixes. Examples for the subword 300 unit set include `interactions', `unofficial', `humiliated', `decency' and `clunker'. We argue that a small unit set allows us to learn prefixes and suffixes more easily. This does not hold for the bigger unit sets, because most of the tokens seen during training are complete words (97\% when using the subword 10k model). When using smaller unit sets, words are more often split into subword tokens. 

\subsection{Comparison to previous work}
We compare our results to previous work using the 300h Switchboard training set in Table \ref{tab:results}. We focus the comparison on CTC models with grapheme-based unit sets. For decoding strategies without using linguistic information, we report gains compared to our previous character based system trained on the same architecture. Highly tuned models which mainly focus on words as their output unit, such as \cite{audhkhasi2017building}, yield better results. However, word-based models do not profit as much from using an LM during decoding. Using LM information during decoding can be an advantage if for example the acoustic training data does not match the test domain. Another advantage of BPE units is that they are able to recognize arbitrary words. When including LM information during decoding, we still improve by more than 2\% compared to the same system using a character unit set, and report slight improvements compared to the character-based model of \cite{zweig2016advances}.

While attention-based methods only show modest improvements when using bigger unit sets \cite{chan2016latent, chiu2017state}, selecting an appropriate unit set for CTC systems seems to be more important. This is also in line with the results from \cite{li2017acoustic}, who report significant gains when switching from characters to words as output units. We argue that it is more difficult for CTC models to learn an implicit LM, which makes it hard to produce the correct spelling of words when only characters are used as output units.

\begin{table}[th]
  \centering
  \caption{Comparison of our results to related work on grapheme-based CTC ASR systems using no LM at all (`No LM') and using a LM during decoding (`LM'). We report the WER on the Switchboard (SW) and Callhome (CH) subsets of Eval2000. \cite{zenkel2017} represents the character baseline for our BPE experiments [ours].}
  \begin{tabular}{ c  c  c  c  c}
  Category & Unit set & SW & CH \\
  \hline
  \hline
  No LM \cite{zenkel2017} & Character & 30.4\% & 44.0\% \\
  No LM [ours] & Subword 10k & 17.8 \% & 29.0\% \\
  No LM \cite{audhkhasi2017building} & Words \& Characters & 14.4\% & 24.0\% \\
  \hline
  LM \cite{zenkel2017} & Character & 17.0\% & 30.2\% \\
  LM [ours] & Subword 300 & 14.7\% & 26.2\% \\
  LM \cite{zweig2016advances} & Characters & 15.1\% & 26.3\% \\
\end{tabular}
\label{tab:results}
\end{table}

Another advantage of BPE-based approaches to create the units is that we can create a unit set of an arbitrary size. Thus, we can adapt the size of our unit set to the size of the training data. Furthermore, we can easily create a number of diverse models which are encouraged by their distinctive unit sets to learn different concepts. We combined the crossword and subword systems decoded with the RNNLM with ROVER \cite{fiscus1997post} using majority voting without any confidence scores.
This yields a word error rate of 11.2\% on the Switchboard test subset, which represents a 3.5\% improvement compared to our best single system.

\section{Conclusions}
\label{sec:conclusions}
In this paper we discussed two different methods to create a unit set for CTC-based speech recognition systems. Our method creates unit sets of any desired size, thus providing a method to conveniently adjust the size of the unit set to the amount of the available training data. 

We believe that this work shows that there is still room for improvement in automatically selecting a unit set for a given dataset. We will continue to improve the automatic selection of units, especially to remedy the drawbacks of crossword models. Using more knowledge about the input sequence to directly benefit from the pronunciation might be a good starting point. 

We argue that the performance of methods like Gram-CTC and Latent Sequence Decomposition could improve when using a BPE unit sets, but also methods relying on less training data usage might be beneficial for systems focusing on a variety of low-resource languages. Also, in situations where the acoustic training data does not match the test domain, one might want to rely more on the information from the language model. Our results suggest that this is possible by specifying a unit set of a smaller size. In the future, we will investigate the use of BPE units in more diverse training and testing scenarios. 


\bibliographystyle{IEEEtran}

\bibliography{mybib}

\end{document}